\UseRawInputEncoding
\documentclass[conference]{IEEEtran}
\IEEEoverridecommandlockouts
\usepackage{amsmath,amssymb,amsfonts}
\usepackage[accsupp]{axessibility} 
\usepackage{graphicx}
\usepackage{amsmath}
\usepackage{amssymb}
\usepackage{booktabs}
\usepackage[pagebackref,breaklinks,colorlinks]{hyperref}
\usepackage{xcolor}
\usepackage{caption}
\usepackage[ruled,vlined]{algorithm2e}
\usepackage{mathtools}
\usepackage{svg}
\usepackage{stackengine}
\usepackage{siunitx}
\usepackage{orcidlink}
\definecolor{reco_orange}{RGB}{255, 191, 128}
\definecolor{restorer_blue}{RGB}{204, 204, 255}
\definecolor{aruco_green}{RGB}{128, 255, 128}
\definecolor{segmentation_gray}{RGB}{204, 204, 204}
\newboolean{long_appendix}
\setboolean{long_appendix}{false}

\usepackage{tikz}
\usepackage{array,multirow,graphicx}
\usetikzlibrary{shapes,arrows, automata}
\usepackage{float}
\usetikzlibrary{scopes}
\usetikzlibrary{backgrounds}
\usetikzlibrary {shadows} 
\pgfdeclarelayer{edgelayer}
\pgfdeclarelayer{nodelayer}
\pgfsetlayers{edgelayer,nodelayer,main}

\usepackage{amsmath}
\usepackage{xcolor}
\usepackage{listings}

\lstdefinestyle{mystyle}{
    backgroundcolor=\color{white},
    commentstyle=\color{green},
    keywordstyle=\color{blue},
    stringstyle=\color{red},
    basicstyle=\ttfamily,
    breaklines=true,
}

\usepackage[capitalize]{cleveref}
\crefname{section}{Sec.}{Secs.}
\Crefname{section}{Section}{Sections}
\Crefname{table}{Table}{Tables}
\crefname{table}{Tab.}{Tabs.}

\usepackage{algpseudocode}
\usepackage{comment}
\usepackage{makecell}
\usepackage{colortbl}
\usepackage[export]{adjustbox}
    
\definecolor{codegray}{gray}{0.9}
\def\BibTeX{{\rm B\kern-.05em{\sc i\kern-.025em b}\kern-.08em
    T\kern-.1667em\lower.7ex\hbox{E}\kern-.125emX}}
\begin{document}

\title{OmniPlantSeg: Species Agnostic 3D Point Cloud Organ Segmentation for  High-Resolution Plant Phenotyping Across Modalities}

\makeatletter
\newcommand{\linebreakand}{%
  \end{@IEEEauthorhalign}
  \hfill\mbox{}\par
  \mbox{}\hfill\begin{@IEEEauthorhalign}
}
\makeatother

\author{\IEEEauthorblockN{Andreas Gilson}
\IEEEauthorblockA{\textit{Fraunhofer Institute for Integrated Circuits (IIS)}\\
Fürth, Germany \\
andreas.gilson@iis.fraunhofer.de}
\and
\IEEEauthorblockN{Lukas Meyer}
\IEEEauthorblockA{\textit{Friedrich-Alexander-Universität  (FAU)}\\
Erlangen-Nürnberg, Germany \\
lukas.meyer@fau.de}
\linebreakand
\IEEEauthorblockN{Oliver Scholz}
\IEEEauthorblockA{\textit{Fraunhofer Institute for Integrated Circuits (IIS)}\\
Fürth, Germany \\
oliver.scholz@iis.fraunhofer.de}
\and
\IEEEauthorblockN{Ute Schmid}
\IEEEauthorblockA{\textit{University of Bamberg}\\
Bamberg, Germany \\
ute.schmid@uni-bamberg.de}
}

\maketitle

\begin{abstract}
Accurate point cloud segmentation for plant organs is crucial for 3D plant phenotyping. Existing solutions are designed problem-specific with a focus on certain plant species or specified sensor-modalities for data acquisition. Furthermore, it is common to use extensive pre-processing and down-sample the plant point clouds to meet hardware or neural network input size requirements. We propose a simple, yet effective algorithm \textit{KD-SS} for sub-sampling of biological point clouds that is agnostic to sensor data and plant species. The main benefit of this approach is that we do not need to down-sample our input data and thus, enable segmentation of the full-resolution point cloud. Combining \textit{KD-SS} with current state-of-the-art segmentation models shows satisfying results evaluated on different modalities such as photogrammetry, laser triangulation and LiDAR for various plant species. We propose \textit{KD-SS} as lightweight resolution-retaining alternative to intensive pre-processing and down-sampling methods for plant organ segmentation regardless of used species and sensor modality.
\end{abstract}


\section{Introduction}
\begin{figure*}[t]
     \centering
     \includegraphics[width=\linewidth]{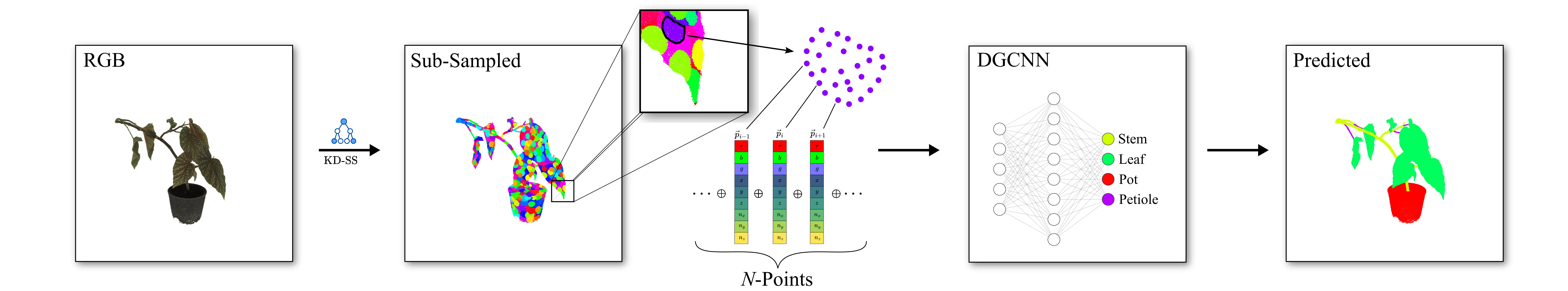}
      \caption{OmniPlantSeg: The raw point cloud is sub-sampled using \textit{KD-SS} with \textit{N}-points per sub-sample retaining the original resolution. The sub-samples with \textit{N} point-wise labels are used as input data for DGCNN. Combining all segmented sub-samples at the end yields the full-resolution point cloud with labels.}
      \label{fig:pipeline}
\end{figure*}
Non-destructive scanning of plants is a key ingredient in the ongoing transformation of agriculture into an increasingly digital industry~\cite{non_destructive_scanning}. Recent improvements in scanning technology and advanced data processing techniques, such as deep learning~\cite{ai_in_agriculture}, shift the data paradigm for phenotyping research from 2D to 3D point clouds~\cite{non_destructive_scanning_pcd}. 
Increasing scanning resolution and fine-grained measurement technology enable the extraction of plant features at a highly detailed scale. However, accurate 3D representation alone is not sufficient for plant phenotyping. In order to extract meaningful information from the scans, reliable plant organ segmentation is required. There are several deep learning architectures available \cite{qi_pointnet_2017-2,DGCNN,li_pointcnn_2018,qu_end--end_2024,zhang_point_2024} and also conventional methods for segmentation \cite{song_comprehensive_2025}. However, reliable segmentation of plant organs remains a challenging topic.

\subsection{Problem Statement and Contribution}
Existing solutions for point cloud segmentation in plant phenotyping are limited by domain restrictions such as plant species or sensor modality \cite{turgut_rosesegnet_2022}. Furthermore, neural network architectures mostly rely on fixed input sizes for 3D segmentation networks, which is usually mitigated by down-sampling of the point clouds and thus immensely reduces the point cloud quality and resolution. An example for this is the workflow of one of the few existing species agnostic approaches PlantNet by Li et. al \cite{li_plantnet_2022} that reduces the point number of each plant to 4096 points. While this is an adequate solution for many use cases, it remains questionable for high-resolution scans with the goal of capturing tiny features and small details. 

To summarize the current problems of the current state of the art in 3D plant segmentation:
\begin{enumerate}
    \item Current segmentation approaches are highly adapted to single domains (e.g. single species or sensor modality).
    \item 3D segmentation networks have limited input sizes and/or high computational requirements.
    \item  Existing solutions rely on down-sampling algorithms or attention mechanisms that reduce informational depth and potentially lead to oversight of small features. 
\end{enumerate}

This paper tackles the problems listed above by making the following contributions:
\begin{enumerate}
\item Showcase a novel and yet simple sub-sampling approach \textit{KD-SS} that splits point clouds of arbitrary size into sub-samples while retaining full resolution.
\item Presenting a domain- and modality-agnostic pre-processing pipeline based on \textit{KD-SS} for accurate 3D point cloud segmentation. 
\item Demonstrate the effectiveness of our \textit{OmniPlantSeg} pipeline using different datasets with multiple plant species and modalities 
and compare it with state-of-the-art benchmarks.
\end{enumerate}


\section{Method: Sub-sampling Algorithm \textit{KD-SS}}
\textit{KD-SS} is an improved adaptation of the \textit{Spherical Sub Sampling} concept proposed by Scholz et al \cite{sphericalsubsampling} that segments point clouds into sub-spheres based on local neighborhoods determined by fixed radii. 
The fixed radius approach of \textit{ Spherical Sub Sampling} distortes the point cloud density distributions, meaning that sub-spheres either contain duplicate data points or remove points from the data. Our approach \textit{KD-SS} is optimized for faster runtime utilizing the KD-tree \cite{bentley_multidimensional_1975,scikit-learn} algorithm that yields sub-samples with varying radius size, while keeping all points of the original point cloud. It works by segmenting datasets into smaller sub-samples, suitable for fixed input size neural networks and edge or consumer grade gpus without heavy computational or VRAM requirements. 

\textit{KD-SS} consists out of the following steps, initialization with a point cloud $D$ and a fixed number of points per resulting sub-sphere $N$, selection of the group of points for the next sub-sample, saving the sub-sample with the desired feature vectors and then, removing the $N$ sub-sampled points from $D$. Fig \ref{fig:KDSS_code} provides pseudo-code of an exemplary implementation of \textit{KD-SS}.
\definecolor{mycolor}{RGB}{139,0,139}
\begin{figure}[ht]
    \centering
    {\scriptsize
\begin{lstlisting} 
(*@\textbf{Input parameters:}@*)
    Point cloud data: (*@\textcolor{mycolor}{\textit{D}} @*) 
    Number of points per sub-sphere: (*@\textcolor{mycolor}{\textit{N}} @*)

(*@\textbf{Variables:} @*)
    Center point of a sub-sphere: (*@\textcolor{mycolor}{\textit{c}} @*)
    Index of already sub-sampled points: (*@\textcolor{mycolor}{{ind-ss}} @*)
    Index of un-sampled points: (*@\textcolor{mycolor}{{remaining-p}} @*)

(*@\textbf{Core-loop:} @*)
while (*@\textcolor{mycolor}{{remaining-p}} @*) > (*@ \textcolor{mycolor}{\textit{N}} @*):
    create KD-Tree of (*@\textcolor{mycolor}{\textit{D}} @*)
    while no duplicate points in (*@\textcolor{mycolor}{{ind-ss}} @*):
        select random (*@\textcolor{mycolor}{\textit{c}}@*) from (*@ \textcolor{mycolor}{{remaining-p}}@*)
        sub-sample = (*@\textcolor{mycolor}{\textit{N}}@*) nearest neighbors of (*@\textcolor{mycolor}{\textit{c}}@*)
        update (*@\textcolor{mycolor}{{ind-ss}} @*)
    update (*@\textcolor{mycolor}{{remaining-p}}@*) by removing (*@\textcolor{mycolor}{{ind-ss}}@*)
    
Save (*@\textcolor{mycolor}{{remaining-p}}@*) as last sub-sample

\end{lstlisting}
}
    \caption{Pseudo-code for \textit{KD-SS} without additional feature generation.}
    \captionsetup{position=bottom}
    \label{fig:KDSS_code}
\end{figure}

The mechanisms for center point selection (line 14 in Fig.\ref{fig:KDSS_code} random-based) can be adapted to alternative approaches, e.g. based on geometry or core-features. Data augmentation or additional feature vectors can be calculated on point- or sub-sample-level (e.g. coordinate normalization of sub-samples). Exemplary descriptors for a biological scan are: $x, y, z$ and $r, g, b$ or $laser$ $intensity$ and $class$. Depending on the use case arbitrary additional per-point dimensions can be added (e.g. point normals, greyscale values, normalized coordinates, etc.). Furthermore, the resulting data samples can also be post-processed depending on the use case e.g. for data augmentation or optional down-sampling.

 After the segmentation is completed and every point receives an additional descriptor: $predicted$ $class$.
 
\section{Experiments}

\begin{table*}[ht]
\centering
\resizebox{0.9\linewidth}{!}{
\begin{tabular}{l|cccccccc}
\multicolumn{8}{c}{\textbf{Datasets overview}}\\
\toprule
\textbf{Dataset} & \textbf{Cherry trees} & \textbf{Wheat field plots} & \multicolumn{2}{c}{ \textbf{Sorghum}} & \multicolumn{3}{c}{ \textbf{Pepper, Rose, Ribes}} \\
\midrule
Source & semi-public \cite{meyer_for5g_2023} & semi-public \cite{virlet_field_2016} & \multicolumn{2}{c}{public \cite{patel_deep_2023}} & \multicolumn{3}{c}{public \cite{mertoglu_planest-3d_2024}} \\
Modality & SfM-MVS & Laser triangulation  & \multicolumn{2}{c}{LiDAR + RGB}& \multicolumn{3}{c}{SfM-MVS} \\
Environment & Outdoor & Outdoor  & \multicolumn{2}{c}{Greenhouse}& \multicolumn{3}{c}{Indoor} \\
Point features & RGB, normals (6) & Intensity (1) & \multicolumn{2}{c}{RGB (3)} & \multicolumn{3}{c}{RGB, normals (6)} \\
Classes & \makecell[c]{ground, trunk, branch, \\ sign, calibration unit, \\ roof structure} & stem, leaf/rest, ear & \multicolumn{2}{c}{stem, leaf, panicle} & \multicolumn{3}{c}{stem, leaf} \\
Data (M points) & 16 trees (95.1) & 3 plots (35.5) & \multicolumn{2}{c}{502 plants (94)}& \multicolumn{3}{c}{34 plants (54.8)} \\
Test set (M points) & 1 tree (4.7 ) & 1 plot (11.3) & \multicolumn{2}{c}{48 plants (8.6)} & \multicolumn{3}{c}{10 plants (13)}\\
\multicolumn{8}{c}{} \\
\multicolumn{8}{c}{\textbf{Segmentation results}} \\
\toprule
Metric & \textbf{Ours} & \textbf{Ours} & \makecell[c]{ \textbf{Authors}\\ (PointNet++)} & \textbf{Ours} & \makecell[c]{ \textbf{Authors}\\ (RoseSegNet)} & \textbf{Ours}  & \makecell[c]{ \textbf{Ours}\\ \textit{shared-weights}}\\
\midrule
Acc              & 97.9 & 76.1 & 92.5 &  94.4 & 98.1 & 94.9 & 95.5 \\
Mean IoU         & 94.3 & 46.2 & 91.6 & 84.9 & 94.2 & 84.5 & 86.3 \\
\end{tabular}
}
\caption{Overview of dataset properties and our semantic segmentation experiments. \textit{OmniPlantSeg} was benchmarked on biological scans of distinct plant species measured with multiple modalities.}
\label{tab:datasetoverview}
\end{table*}

\subsection{Datasets}
\begin{figure}
    \centering
    \includegraphics[width=0.7\linewidth]{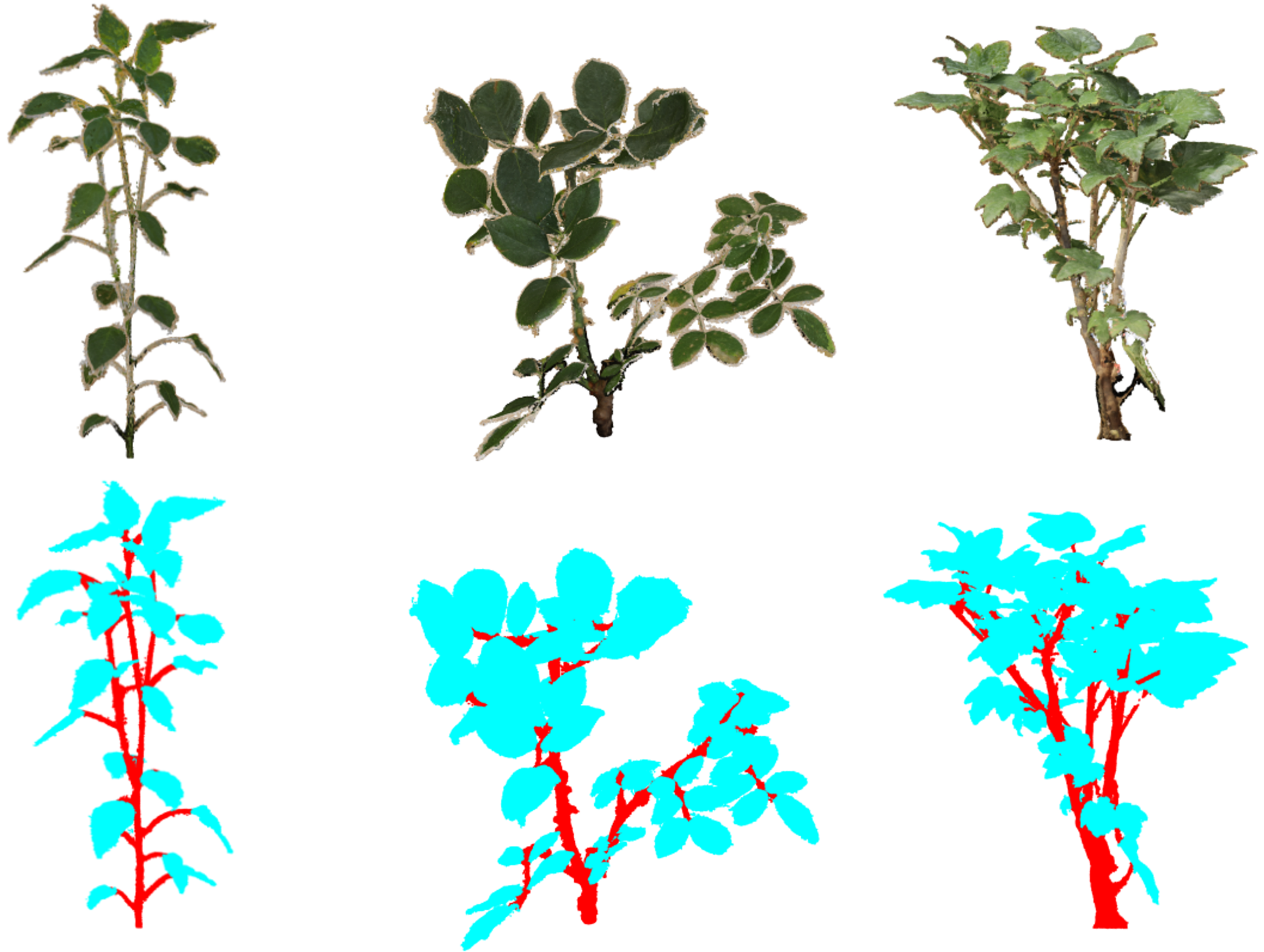}
    \caption{Pepper (left), rose (middle) and ribes (right) data in RGB (top) and annotated (bottom) with classes \textit{stem} (red) and \textit{leaf} (cyan) from \cite{mertoglu_planest-3d_2024}.}
    \label{fig:planest_data}
\end{figure}
\begin{figure}
    \centering
    \includegraphics[width=0.7\linewidth]{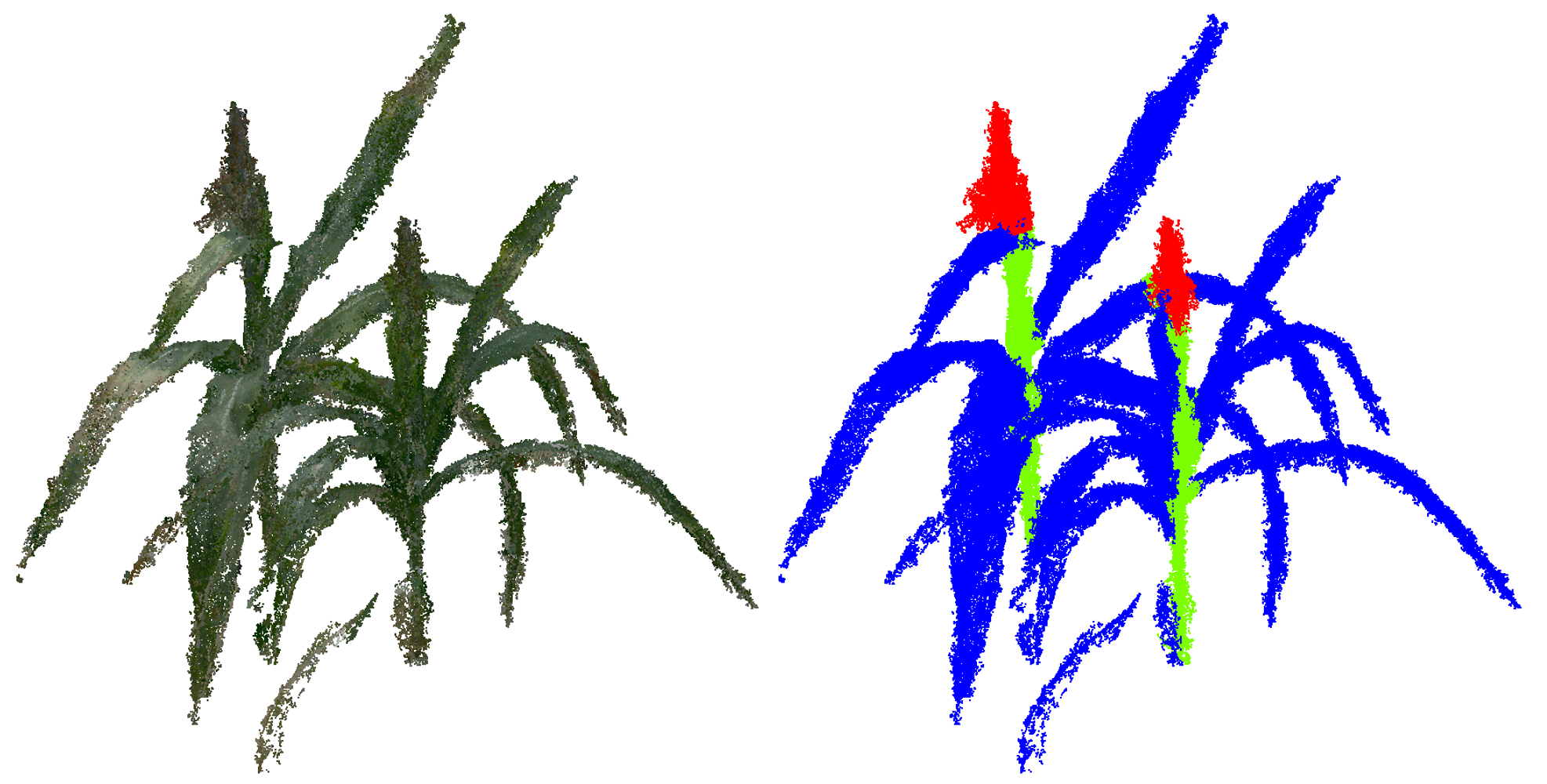}
    \caption{Sorghum \cite{patel_deep_2023} in RGB (left) and annotated (right) with three classes: \textit{stem} (green), \textit{leaf} (blue) and \textit{panicle} (red).}
    \label{fig:sorghum_data}
\end{figure}
\begin{figure}
    \centering
    \includegraphics[width=0.7\linewidth]{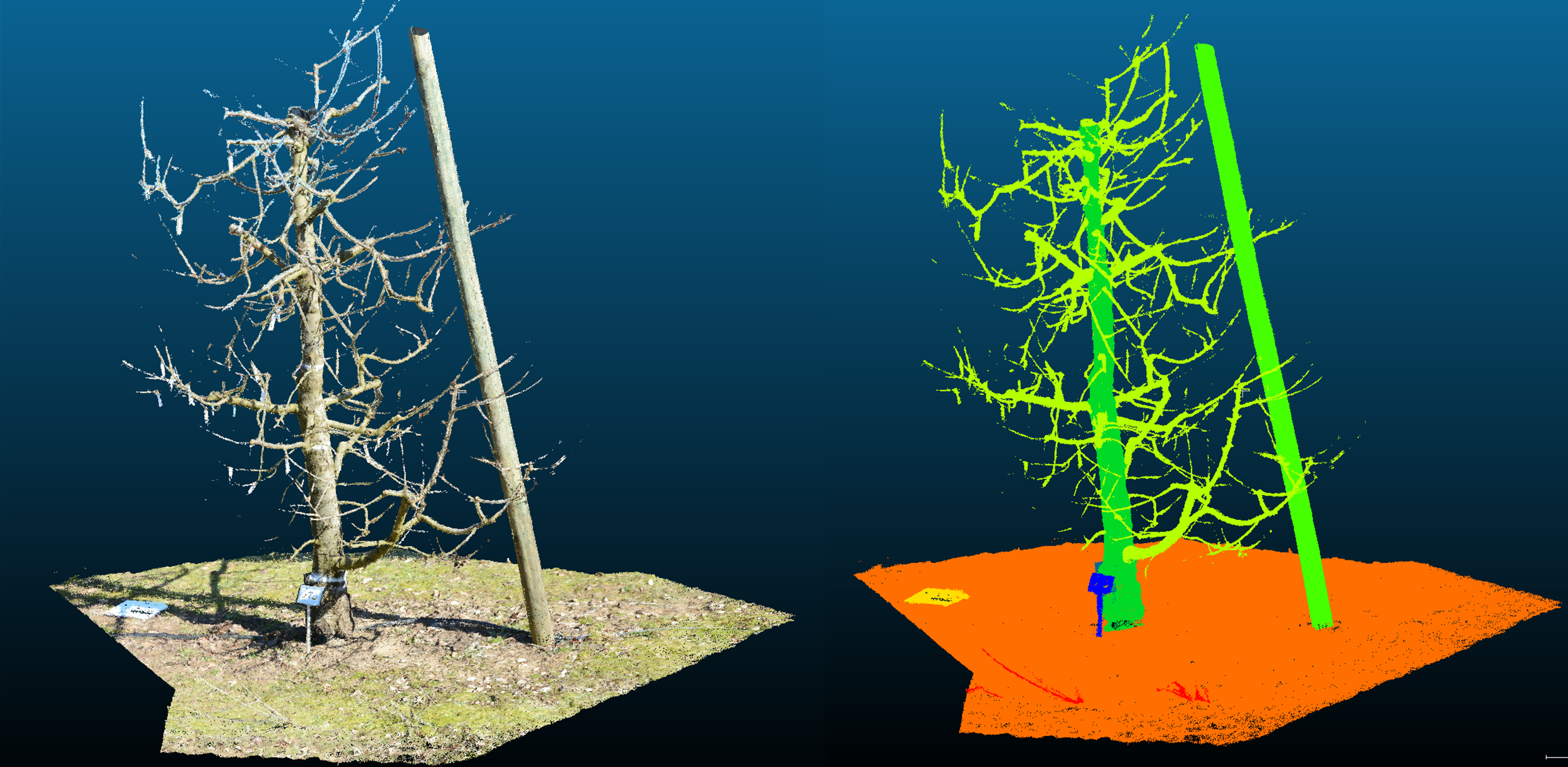}
    \caption{Cherry tree \cite{meyer_for5g_2023} in RGB (left) and annotated (right) with multiple color coded classes.}
    \label{fig:forg_data}
\end{figure}
\begin{figure}
    \centering
    \includegraphics[width=0.7\linewidth]{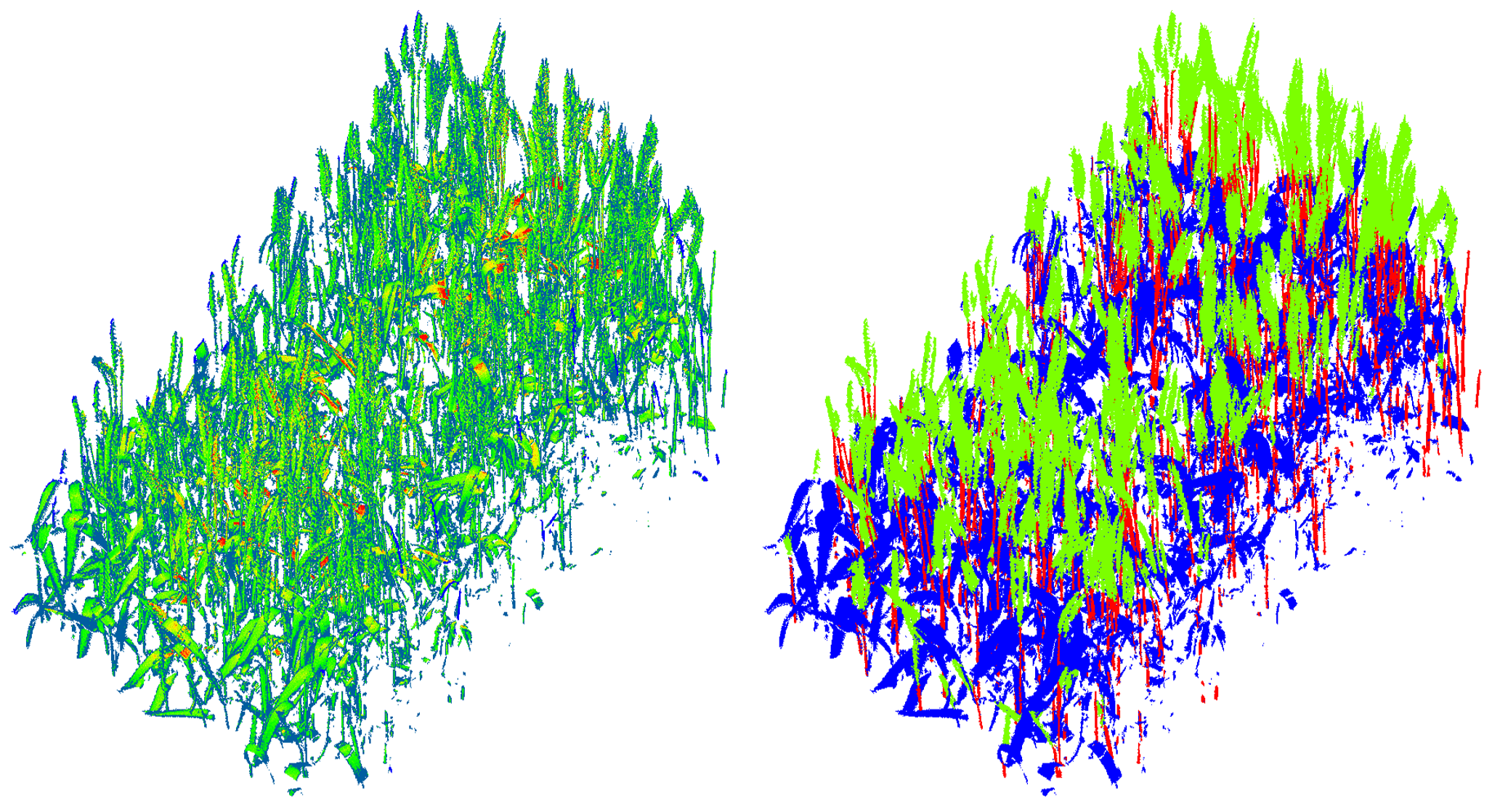}
    \caption{Wheat field plot \cite{virlet_field_2016} with colorized intensity values (left) and annotated (right) with classes \textit{stem} (red), \textit{ear} (green) and \textit{leaf/rest} (blue).}
    \label{fig:wheat_data}
\end{figure}

The \textit{OmniPlantSeg} pipeline was tested on diverse datasets to ensure compatibility with various sensor modalities and distinct plant species. There is no commonly used benchmark dataset for high-resolution plant point clouds. Thus, the experiments focus on datasets that are either public or semi-public, meaning that the authors provide the datasets when asked directly. A comprehensive overview of the datasets used is given in table \ref{tab:datasetoverview}. The first public dataset with benchmark values to compare to is called PLANesT-3D \cite{mertoglu_planest-3d_2024} and consists of 3 species (Pepper, Rose and Ribes) split in two classes \textit{stem} and \textit{leaf} (Fig.\ref{fig:planest_data}). 
The second public dataset consists of Sorghum plants scanned with LiDAR, colored in post-processed \cite{patel_deep_2023} and the three classes \textit{stem}, \textit{leaf} and \textit{panicle} (Fig.\ref{fig:sorghum_data}). The semi-public data of wheat field plots scans were measured with laser triangulation using two 3D lasers in a specialized scanning device \cite{virlet_field_2016}. The three scans are from differently wheat varieties that have been annotated manually into \textit{stem}, \textit{leaf/rest}, \textit{ear} and have single-channel intensity values per point (Fig.\ref{fig:wheat_data}). The last dataset used are semi-public  cherry tree scans in vegetation dormancy that were reconstructed based on RGB images using SfM-MVS \cite{meyer_for5g_2023} (Fig.\ref{fig:forg_data}). In contrast to the PLANesT-3D data the scans are from an outdoor setting and have 6 target classes (\textit{ground}, \textit{trunk},
\textit{branch}, \textit{sign}, \textit{calibration unit}, and \textit{roof structure}). All visualizations of point clouds were done using the open software CloudCompare \cite{cc}.

For public datasets we followed the the train-test split of the original authors. A fixed split of randomly chosen 15 percent of the training data was used for validation during training. For the semi-public datasets no benchmarks were available, thus we separated test data as holdout sets as displayed in Tab. \ref{tab:datasetoverview} and sub-sampled the remaining data before splitting randomly 90/10 into train/validation.


\subsection{Semantic Segmentation: DGCNN}

The proposed \textit{KD-SS}  outputs an arbitrary feature depth for every point, which has to be considered when adapting the neural network architecture for semantic segmentation. Our experiments are based on Dynamic Graph Convolultion (DGCNN) \cite{DGCNN}  implementation in PyTorchGeomentric \cite{PyTorchGeometric}. DGCNN proved to work well in combination with \textit{KD-SS} and outperformed competitors like PointNet++ \cite{qi_pointnet_2017-1} in previous experiments \cite{li_comparative_2023}. DGCNN was setup as default with \textit{k=30}, \textit{dropout=0.3} and active batch normalization. Otherwise, it was only adapted to match the target classes and data feature channels of the respective dataset. 
All runs were trained on \textit{negative-log-likelihood-loss} and \textit{Adam} optimizer. The initial \textit{learning-rate=0.0081} was scheduled to be reduced on plateau. As evaluation metrics we used \textit{Precision}, \textit{Recall}, \textit{Intersection over Union (IoU)}, and \textit{Accuracy}, reported either per class $C$ or as mean averaging the values over all classes. For example, accuracy ($\frac{\text{correct predictions}}{\text{total predictions}}$) and mean accuracy ($\frac{1}{C} \sum_{c=1}^{C} \text{Accuracy}_c$). For a comprehensive overview of the other metrics, refer to Li et al.~\cite{li_comparative_2023}.
Adjusting the loss function with normalized class weights based on the class distribution on the training dataset was only done for the six-class cherry tree dataset. We ran our experiments on a cluster with \textit{Nvidia RTX A5000} and \textit{A6000} GPUs and tested all datasets using a single GPU workstation equipped with a \textit{Nvidia GeForce RTX 2080 Super 8GB}, to verify \textit{OmniPlantSeg} being compatible with consumer grade and edge hardware like \textit{ Nvidia Jetson boards}. Training convergence times ranged from 5 to 19 hours, depending on the batch sizes and dataset. Model inference for a single plant point cloud takes less than 1 minute. The only change to the pipeline required to adapt the \textit{OmniPlantSeg} workflow on less powerful GPUs is reduction of batch size. Although, lower batch sizes increased model convergence time, we found no general impact on final training metrics or prediction quality with the tested batch sizes of 8,16,32 and 64.

\section{Results}
The results of our experiments on the PLANesT-3D dataset are compared to the work of the original authors in Tab. \ref{tab:planest_results}. We trained individual networks for each species and one model on all combined training datasets that we call \textit{shared weights}. The \textit{shared weights} run was evaluated using the species-wise test datasets as done by the authors and our single-species models. Notably, our method achieves the highest mean IoU in the pepper dataset, indicating robust segmentation capabilities even without extensive pre-processing. While our shared-weights approach shows a slight decrease in performance compared to species-specific weights, it still maintains reasonable accuracy, suggesting the versatility across different plant types. In particular, leaf segmentation works consistently across all datasets. However, stem segmentation performance indicates room for improvement, especially in datasets where stems are underrepresented (e.g., ribes). 

Sorghum segmentation results are showcased in Tab. \ref{tab:sorghum_results}. Our method, while trailing behind in recall, demonstrates the best accuracy and thus, feasibility of segmentation without heavy pre-processing. Despite engaging with full-resolution LiDAR data with relatively coarse resolution, our pipeline yields competitive results. While there is room for improvement for the class \textit{stem}, the segmentation results for leaves and the new class \textit{panicle} are robust.
The wheat data was provided with 1-channel intensity values representing the laser reflectance instead of instead of 3 color channels, resulting in a feature vector length of 7, since we added coordinate normalization on sub-sample level as additional features. The accuracy of segmentation across wheat classes is 0.76, with notably high precision and recall for the \textit{leaf/rest} class. Specifically, the \textit{leaf/rest} class achieved a precision of 0.72 and a recall of 0.94, resulting in a favorable F1-score of 0.81. The \textit{ear} class demonstrated similar F1 performance, with precision and recall of 0.85 and 0.77, respectively. However, the \textit{stem} class presented significant challenges, evidenced by a precision of 0.82 but a recall of only 0.02, leading to a low F1-score of 0.05.

The cherry tree dataset is challenging because the increased number of target classes with a significant imbalance. The SfM-MVS point clouds has RGB values and normals creating a feature vector with length 9 for each point. The segmentation performance of an overall \textit{Accuracy = 0.98}, and total \textit{IoU = 0.96} which was slightly worse when averaged over all classes \textit{MIoU = 0.94}. Class \textit{ground} had the best performance and \textit{sign} the worst (Tab.\ref{tab:datasetoverview}).

\begin{figure}
    \centering
    \includegraphics[width=0.7\linewidth]{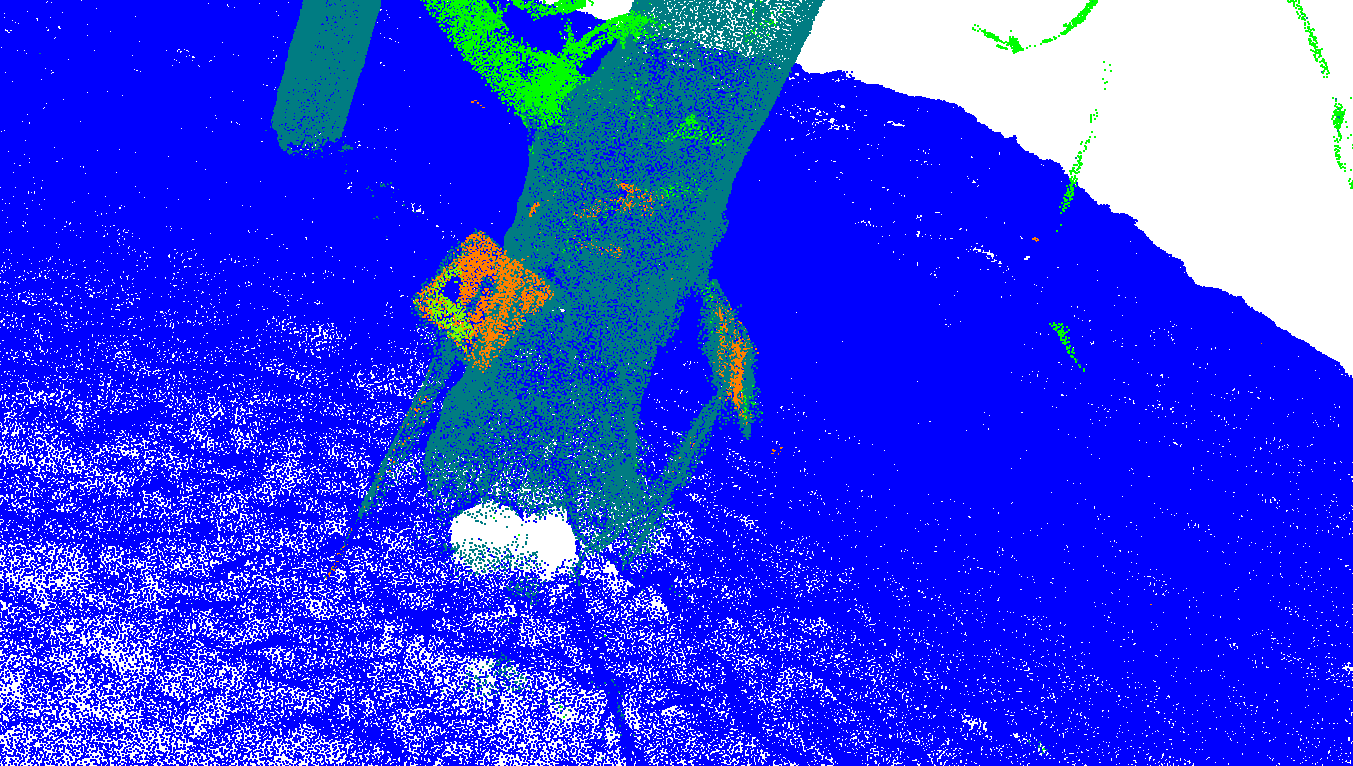}
    \caption{Closeup on segmentation results of cherry tree with focus on underrepresented class \textit{sign} and \textit{trunk} (dark green), \textit{ground} (blue) and \textit{branches} (bright green) in the background.}
    \label{fig:for5g_sign}
\end{figure}

\begin{figure}
    \centering
    \includegraphics[width=0.7\linewidth]{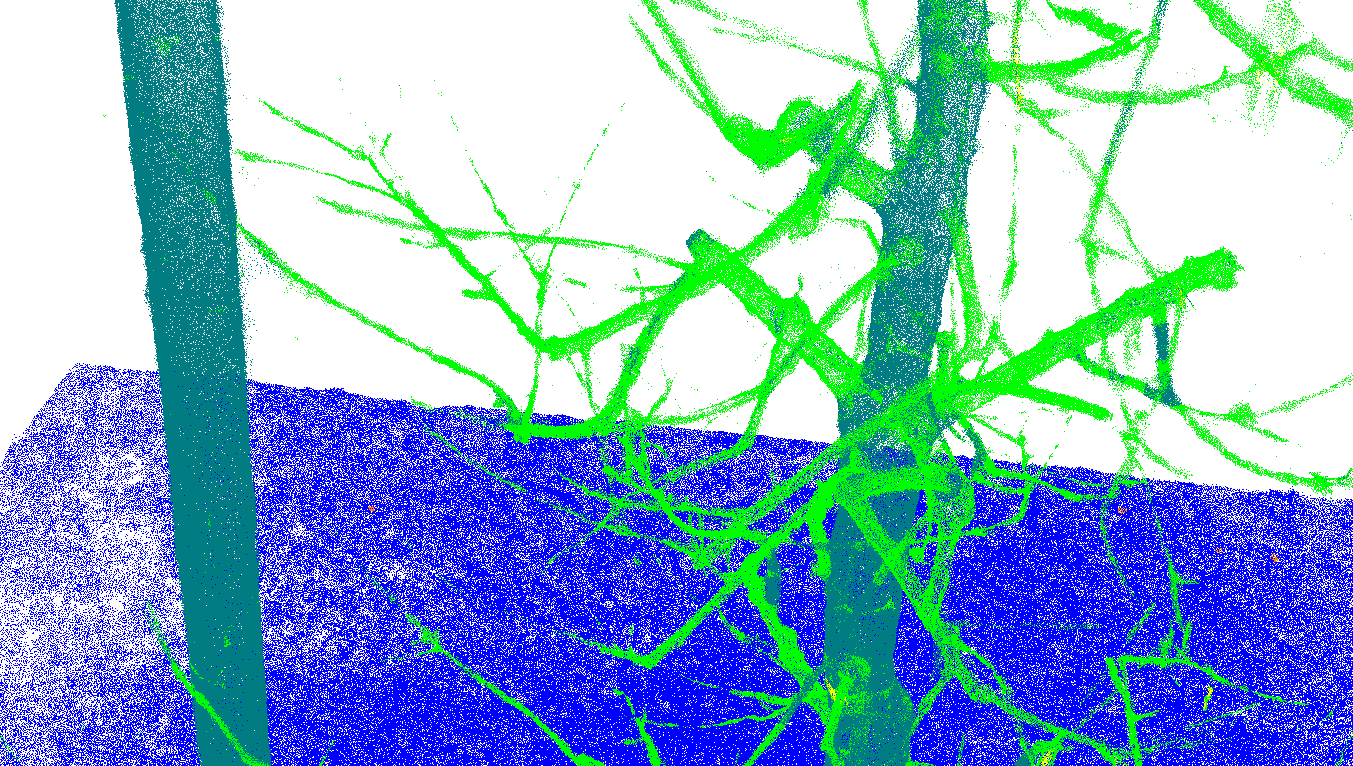}
    \caption{Closeup on segmentation results of cherry tree with focus on branching structure and target classes \textit{trunk} (dark green), \textit{ground} (blue) and \textit{branches} (bright green).}
    \label{fig:forg_branch}
\end{figure}

\begin{table}[h]
\centering
\resizebox{\columnwidth}{!}{
\begin{tabular}{l|c|c|c|c|c}
\multicolumn{6}{c}{\textbf{PEPPER}} \\
\toprule
\textbf{Metric} & PointNet++ & RoseSegNet & SP-LSCnet & \textbf{Ours} & \makecell[c]{ \textbf{Ours}\\ \textit{shared-weights}} \\
\midrule
Precision - Stem & 96.5 & 97.0 & \cellcolor{green!25}97.6 & 96.7 & 96.7\\
Recall - Stem    & 94.6 & 96.1 & 94.5 & 96.3 & 94.8\\
IoU - Stem       & 91.5 & \cellcolor{green!25}93.2 & 92.4 & \cellcolor{green!25} 93.2 & 91.8\\
Precision - Leaf & 98.3 & \cellcolor{green!25}98.7 & 98.3 & 93.2 & 98.5\\
Recall - Leaf    & 98.9 & 99.0 & \cellcolor{green!25} 99.3 & 99.0 & 99.0\\
IoU - Leaf       & 97.2 & 97.8 & 97.6 & \cellcolor{green!25} 98.0 & 97.6\\
Acc              & 97.9 & 98.3 & 98.1 & \cellcolor{green!25} 98.4 & 98.1\\
MIoU             & 94.3 & 95.5 & 95.0 & \cellcolor{green!25} 95.6 & 94.7
\\
\\
\multicolumn{6}{c}{\textbf{ROSE}} \\
\toprule
\textbf{Metric} & PointNet++ & RoseSegNet & SP-LSCnet & \textbf{Ours} & \makecell[c]{ \textbf{Ours}\\ \textit{shared-weights}} \\
\midrule
Precision - Stem & 92.4 & 94.8 & \cellcolor{green!25}96.4 & 88.8 & 86.0\\
Recall - Stem    & 89.7 & \cellcolor{green!25}91.4 & 85.6 & 78.2 & 78.5\\
IoU - Stem       & 83.5 & \cellcolor{green!25}87.1 & 83.0 & 71.2 & 69.6\\
Precision - Leaf & 97.6 & \cellcolor{green!25}98.0 & 97.2 & 95.9 & 95.9\\
Recall - Leaf    & 98.3 & 98.8 &\cellcolor{green!25} 99.4 & 98.1 & 97.5\\
IoU - Leaf       & 96.0 & \cellcolor{green!25}96.9 & 96.6 & 94.1 & 93.6\\
Acc              & 96.7 & \cellcolor{green!25}97.4 & 97.1 & 94.9 & 94.4\\
MIoU             & 89.8 & \cellcolor{green!25}92.0 & 89.8 & 82.7 & 81.6
\\
\\
\multicolumn{6}{c}{\textbf{RIBES}} \\
\toprule
\textbf{Metric} & PointNet++ & RoseSegNet & SP-LSCnet & \textbf{Ours} & \makecell[c]{ \textbf{Ours}\\ \textit{shared-weights}} \\
\midrule
Precision - Stem & 95.8 & 95.9 & \cellcolor{green!25}96.7 & 77.8 & 80.8\\
Recall - Stem    & 94.0 & \cellcolor{green!25}95.5 & 94.0 & 72.2 & 86.8\\
IoU - Stem       & 90.3 & \cellcolor{green!25}91.7 & 91.1 & 59.9 & 72.0\\
Precision - Leaf & 98.9 & \cellcolor{green!25}99.2 & 98.5 & 94.2 & 97.1\\
Recall - Leaf    & \cellcolor{green!25}99.2 & \cellcolor{green!25}99.2 & \cellcolor{green!25}99.2 & 95.6 & 95.6\\
IoU - Leaf       & 98.1 & \cellcolor{green!25}98.4 & 97.8 & 90.3 & 93.0\\
Acc              & 98.4 & \cellcolor{green!25}98.6 & 98.3 & 91.5 & 94.1\\
MIoU             & 94.2 & \cellcolor{green!25}95.1 & 94.5 & 75.1 & 82.5
\end{tabular}%
}
\caption{Segmentation performance comparison of the PLANesT-3D dataset \cite{mertoglu_planest-3d_2024}.}
\label{tab:planest_results}
\end{table}

\begin{table}[h]
\centering
\resizebox{\columnwidth}{!}{%
\begin{tabular}{l|c|c|c|c|c}
\multicolumn{6}{c}{\textbf{Sorghum test set performance compared with best runs from Patel et al. \cite{patel_deep_2023}}} \\
\toprule
\textbf{Metric} & PointNet & PointNet++ & PointCNN & DGCNN & \textbf{Ours} \\
\midrule
Mean precision & 85.8 & \cellcolor{green!25}94.8 & 92.4 & 89.4 & 92.3\\
Mean recall & 94.6 & \cellcolor{green!25}95.3 & 94.8 & 94.2 & 90.9 \\
Mean Acc & 83.2 & 92.5 & 88.3 & 86.7 & \cellcolor{green!25}96.3 \\
Mean IoU & 81.5 & \cellcolor{green!25}91.6 & 87.3 & 85.3 & 84.9
\end{tabular}%
}
\caption{Segmentation performance comparison on the Sorghum dataset \cite{patel_deep_2023}.}
\label{tab:sorghum_results}
\end{table}

\section{Discussion}
The PLANesT-3D authors rely on intensive pre-processing using voxel-based sub-sampling and a super-point procedure for their experiments. In contrast to \textit{KD-SS} this reduces the effective resolution. This is reflected in the metrics, where the combination of this pre-processing and RoseSegNet \cite{turgut_rosesegnet_2022} led to superior segmentation results. Despite the fact that an increased point number increases the task difficulty, \textit{OmniPlantSeg} demonstrates competitive results against established approaches. The results of the individual and \textit{shared weights} models underline the effectiveness of the \textit{KD-SS} algorithm while maintaining input resolution. However, RoseSegNet seems to outperform DGCNN for the ribes and rose data, suggesting that it remains a better choice for datasets where the \textit{stem} class is underrepresented and extensive pre-processing is not problematic. The results of our \textit{shared-weights} model demonstrate that single models can generalize well over different species, making the development of a foundation model for this task a promising follow-up research topic.

For the Sorghum dataset the spatial distribution of the plants proves to be an additional challenge. The authors of the dataset down-sampled the point clouds and normalized the coordinates to mitigate this problem. Our experiments were performed on full-resolution point clouds without down-sampling or altering of positions and coordinates, which explains the performance differences displayed in Tab. \ref{tab:sorghum_results}. We added the coordinates normalized per sub-sphere as additional features to the network input. Since this is a trivial extension of the standard \textit{KD-SS}, this could eliminate the need for spatial normalization in pre-processing. These results underscore the importance of balancing input resolution with segmentation accuracy, suggesting further optimization of our pipeline may be necessary to leverage the benefits of \textit{KD-SS} fully. Nonetheless, this approach remains a promising alternative for scenarios where preserving data integrity is crucial, despite the observed trade-offs in segmentation metrics.

Spatial distribution also was a factor for the laser triangulation wheat field point clouds, which were challenging to segment. This can be partly explained by the experimental design of the data split. Each wheat plot scan consists out of distinct wheat strain breeds with different spatial positions and noticeable differences in the data. Since we trained and validated on two field plots but tested on a third, our setup represents a realistic use case with the model being forced to generalize within a slight domain shift. The \textit{stem} class had notably low recall and IoU, while the \textit{leaf/rest} and \textit{ear} classes reached competitive precision and recall.

We explain the above average segmentation performance on the SfM-MVS cherry tree data with the distinctiveness of the target classes. The class \textit{ground} for example is spatially fixed to the bottom of the cloud making it a trivial class to learn. The stem without branches next to the tree was assigned to the same target class as the \textit{trunk}, which simplifies the segmentation task. The most difficulties arose during segmentation of vastly underrepresented classes like \textit{sign} (Fig.\ref{fig:for5g_sign}). The branching structure of the tree was only partially reconstructed, making a lossless resolution-retaining approach for segmentation particularly important as displayed in Fig.\ref{fig:forg_branch}. Follow-up processes such as tree topology reconstruction methods rely on high-resolution accurate segmentation results \cite{meyer_cherrypicker_2023}.

\section{Conclusion}
We presented the novel \textit{KD-SS} sub-sampling algorithm that was combined with the DGCNN point cloud segmentation neural network in our pipeline called \textit{OmniPlantSeg} for organ segmentation of high-resolution point clouds in the biological domain. Our experiments benchmarked \textit{OmniPlantSeg} on public and semi-public datasets of six different plant species captured with three distinct techniques and sensor modalities in out- and indoor settings.
OmniPlantSeg showcased species-agnostic segmentation across various modalities. On the PLANesT-3D dataset, OmniPlantSeg achieved a mean IoU of 0.943 for the pepper species, demonstrating competitive performance against existing methods like RoseSegNet \cite{turgut_rosesegnet_2022} and SP-LSCnet \cite{mertoglu_planest-3d_2024}. Notably, the shared weights model maintained reasonable accuracy, indicating potential for generalization among different species. Despite challenges in stem segmentation, particularly in datasets with underrepresented stems, our approach worked well for leaf segmentation, consistently achieving high precision and recall.
For future research, we plan the evaluation of \textit{KD-SS} in combination with other architectures, especially those with instance segmentation capabilities. Other follow-up contributions could be the curation of one benchmark dataset combining all currently available plant point cloud datasets and further pursuing our \textit{shared-weights} attempt of training plant organ foundation models. The research community would also benefit from additional high-resolution plant point clouds created by state-of-the-art computer vision methods like NeRFs \cite{nerfstudio} or Gaussian Splatting \cite{kerbl_3d_2023} that become increasingly relevant in the agricultural domain \cite{meyer_fruitnerf_2024,chopra_agrinerf_2024,meyer_fruitnerf_2025,meyer_multi-spectral_2025}.

\section{Acknowledgements}
The authors express their gratitude to the research groups who provided the invaluable datasets \cite{mertoglu_planest-3d_2024,patel_deep_2023,virlet_field_2016,meyer_for5g_2023} that made this study possible. This project is funded by the 5G innovation program of the German Federal Ministry for Digital and Transport under the funding code 165GU103B.

{\small
\bibliographystyle{IEEEtran}
\bibliography{references,library}
}
\end{document}